# Optimized Learning for X-Ray Image Classification for Multi-Class Disease Diagnoses with Accelerated Computing Strategies

Sebastián A. Cruz Romero, Ivanelyz Rivera de Jesús, Dariana J. Troche Quiñones
Computer Science and Engineering Department
University of Puerto Rico at Mayagüez
Mayagüez, Puerto Rico

*Abstract*— X-ray image-based disease diagnosis lies in ensuring the precision of identifying afflictions within the sample, a task fraught with challenges stemming from the occurrence of false positives and false negatives. False positives introduce the risk of erroneously identifying non-existent conditions, leading to misdiagnosis and a decline in patient care quality. Conversely, false negatives pose the threat of overlooking genuine abnormalities, potentially causing delays in treatment and interventions, thereby resulting in adverse patient outcomes. The urgency to overcome these challenges compels ongoing efforts to elevate the precision and reliability of X-ray image analysis algorithms within the computational framework. This study introduces modified pre-trained ResNet models tailored for multi-class disease diagnosis of X-ray images, incorporating advanced optimization strategies to reduce the execution runtime of training and inference tasks. The primary objective is to achieve tangible performance improvements through accelerated implementations of PyTorch, CUDA, Mixed-Precision Training, and Learning Rate Scheduler. While outcomes demonstrate substantial improvements in execution runtimes between normal training and CUDA-accelerated training, negligible differences emerge between various training optimization modalities. This research marks a significant advancement in optimizing computational approaches to reduce training execution time for larger models. Additionally, we explore the potential of effective parallel data processing using MPI4Py for the distribution of gradient descent optimization across multiple nodes and leverage multiprocessing to expedite data preprocessing for larger datasets.

1. INTRODUCTION

*A. Problem Statement*

The current landscape of X-ray image-based disease diagnosis and multi-class image classification models confronts a significant challenge pertaining to the execution time of training these sophisticated models. As the demand for accurate and timely diagnosis increases, the computational burden imposed by intricate multi-class classification models designed for disease identification becomes a bottleneck. As increased variability and large datasets potentially increases inaccurate results, the execution time also remains a notable concern, impacting the speed at which these models can be trained and deployed for real-world applications.

To address these issues and unlock the potential of different training strategies we explore methods within the PyTorch framework and complimentary modules as there is a demand for an innovative computational solution that not only may enhance accuracy but also streamlines and accelerates the image classification process, ensuring versatility and efficiency in diverse research settings.

*B. Significance of the Problem*

The prolonged execution time during training hinders the seamless integration of these models into clinical workflows, where swift and accurate diagnoses are crucial for timely medical interventions. This delay not only affects the efficiency of healthcare processes but also poses challenges in managing larger datasets inherent in medical imaging. As the scale and complexity of these multi-class image classification models grow, the need for optimizing execution time becomes paramount to ensure their practical utility in the fast-paced clinical environment. Addressing the challenge of prolonged training execution time is pivotal to unlocking the full potential of these advanced models, facilitating their widespread adoption, and ultimately improving the speed and accuracy of disease diagnosis through X-ray images.

*C. Relevant Background*

The theme of data variability highlights the challenges of working with medical images, addressing issues such as handling diverse image dimensions and managing redundant background information. [7] Moreover, the primary emphasis on the prolonged execution time during training in the text aligns with broader concerns about computational efficiency. The interpretation of medical data predominantly relies on the expertise of medical professionals. However, when it comes to the analysis of medical images, this process faces inherent limitations due to its subjective nature and the complexity of the images. Significant variations can exist between different experts, and the demanding workload often leads to fatigue. [6] Compounding this challenge is the increasing pressure on models to deliver both fast and accurate diagnoses. As models strive to meet the dual demand for speed and precision, their efficiency is compromised, contributing to a slowdown in the interpretation process. These are some challenges in medical image processing regarding current computational models complexities, dataset high-dimensionality issues, and the imperative to optimize model training for practical clinical applications.

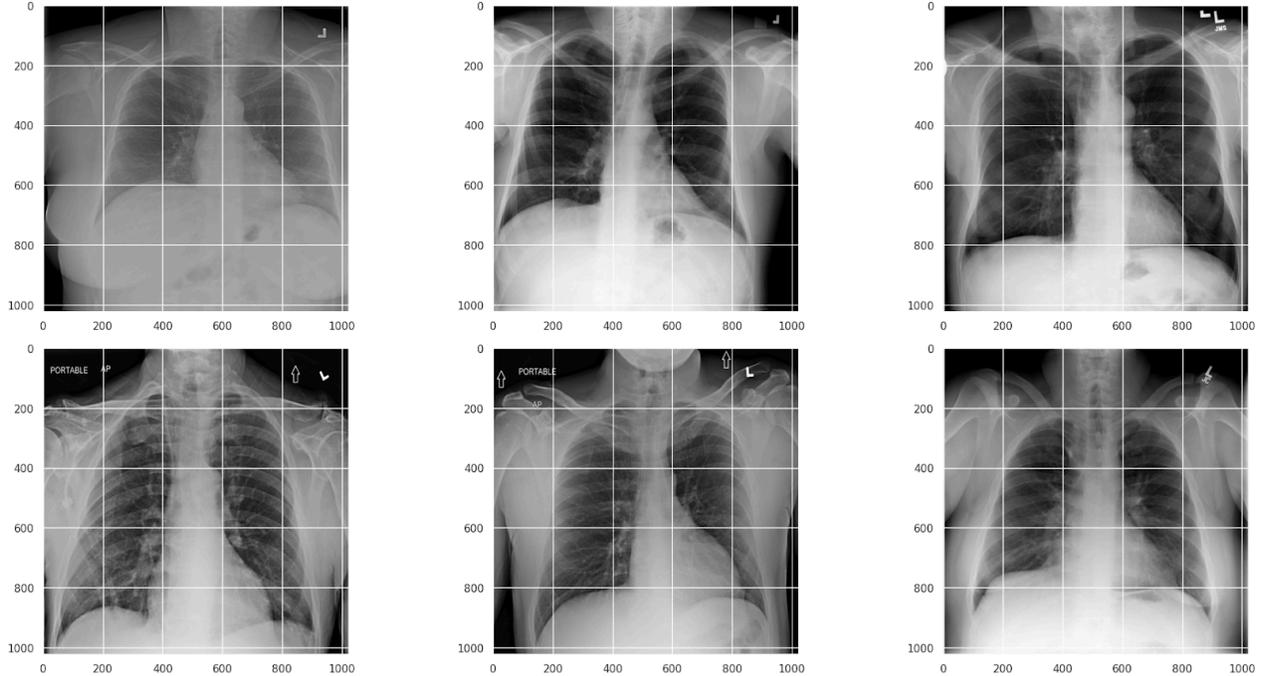

*Figure 1: Random selection of images from the ChestX-Ray8 Dataset. RGB images are transformed to grayscale images and are of width and height 1024x1024.*

2. METHODS

*D. Dataset*

In this study we work with chest x-ray images obtained from the public ChestX-Ray8 Dataset. [9] We use a small subset of data that contains 1000 image samples and two csv records split into training and testing patient IDs. Our data contained 14 classes: Atelectasis, Cardiomegaly, Consolidation, Edema, Effusion, Emphysema, Fibrosis, Hernia, Infiltration, Mass, Nodule, Pleural Thickening, Pneumonia, and Pneumothorax.

Furthermore, we visualize our data to familiarize with the image contents of our dataset. We transform our images into one single channel obtaining a grayscale image with a pixel width and height of 1024x1024 shown in **Figure 1**. We evaluate the pixel intensities of a single image and the pixel value distribution to determine if an image has the correct exposure, overexposed, underexposed, or with an abnormal image crop size in **Figures 2**. Before we feed our data to our model wse standardize our images to adjust our image data to a mean of 0 and a standard deviation of 1 by replacing each pixel value in the standardized image with a value calculated by subtracting the mean and dividing by the standard deviation. Afterwards, we transform our image matrices to tensor values to input our data.

*E. Framework and Tools*

In the context of data exploration and analysis, the code leverages various operations from numpy and pandas. It explores and analyzes the provided medical image dataset, printing information about the number of patient IDs, the total number of samples, and the distribution of positive labels for each medical condition.

In terms of dataset preprocessing, the code utilizes the os and shutil libraries. It includes a section for creating subfolders to organize medical images based on their split (train, validation, test). The script then uses shutil.copy to copy images from a general folder to their respective subfolders, a critical step for organizing the dataset for machine learning model input. For image plotting and analysis, the code utilizes matplotlib and numpy. The plot_images function uses matplotlib to visualize a set of randomly selected medical images, iterating through the images, reading and displaying them, and providing an option for grayscale visualization. The analyze_single_image function performs a detailed analysis of a single image, displaying the raw chest X-ray image and its pixel intensity distribution using matplotlib. The plot_image_hist function is responsible for plotting the histogram of pixel intensities for an image. The timed_execution function relies on the *time* library to measure and print the elapsed time between two specified time points, aiding in performance evaluation.

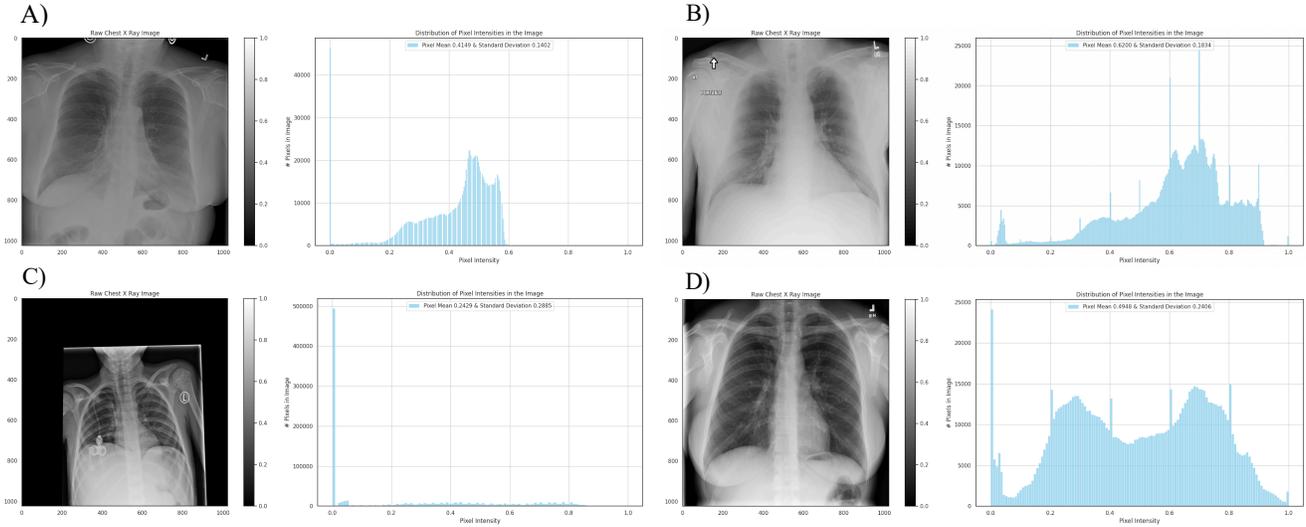

*Figure 2: Pixel intensity and distribution of single images. From left to right: A) overexposed image sample has a pixel mean of 0.4149 and standard deviation of 0.1402, B) underexposed image sample has a pixel mean of 0.6200 and standard deviation of 0.1834 D) padded image sample has a pixel mean of 0.2428 and standard deviation of 0.2885 D) correct exposure image sample has a pixel mean of 0.4948 and standard deviation of 0.2406*

*F. Model Definition and Data Loading*

We create a custom class named CustomDataset to handle the loading and preprocessing of images with their respective labels. It takes parameters such as the root directory, dataset metadata, subset information, and an optional transformation function for image processing. The class includes methods for determining the dataset length and retrieving individual items, with optional image transformations applied during the process.

Moving on to the model definition, a pre-trained ResNet18 architecture is employed for both the normal and CUDA-accelerated models. In the normal model variant, the architecture is instantiated without GPU acceleration. The ResNet18 model is modified to suit the specific classification task, involving the removal of the fully connected layers and the addition of a new fully connected layer tailored for the desired number of output classes (14 in this case).

In the CUDA-accelerated model, the code checks for the availability of a CUDA-compatible GPU and assigns the model to the GPU device if present. This allows for faster training and inference times by leveraging GPU acceleration. The subsequent sections detail the training and testing loops for both models, involving the use of BCEWithLogitsLoss as the loss function, the Adam optimizer, and the application of the defined CustomDataset for loading and preprocessing image data. The training loops iterate over a specified number of epochs, recording training times and losses, while the testing loops evaluate the trained models on a separate test dataset, calculating accuracy by comparing predicted outputs to ground truth labels. Overall, the code provides a comprehensive pipeline for training and evaluating a medical image classification model, demonstrating the flexibility of the CustomDataset class and the adaptability of the ResNet18 architecture for both CPU and GPU execution.

*G. Accelerated Training Strategies*

After comparing the runtime execution of normal training and CUDA-accelerated training, we evaluate different training optimization strategies to evaluate the runtime execution and loss convergence. We evaluated all models with the same image transformations of resizing, tensor conversion, and standardization, and hyperparameters like a batch size of 32, an initial learning rate of 0.001, and over 25 epochs. Moreover, we used an Adam optimizer and Binary Cross Entropy with Logits loss function to evaluate the accuracy of our predictions against our labels.

The first model, referred to as the "Normal Model," utilizes a pre-trained ResNet18 architecture for classifying X-ray images. The second model, denoted as the "CUDA-Accelerated Model," follows a similar structure to the normal model but is designed to leverage GPU acceleration. The device (CUDA or CPU) is selected based on the availability of a compatible GPU. Since we are using Google Colab we used a Tesla T4 GPU in this instance. The training loop includes additional steps to move data and model parameters to the GPU, resulting in faster execution times. The testing process evaluates the model on a separate test dataset, calculating accuracy by comparing predicted outputs to ground truth labels. The third model is based on the second model with added learning rate scheduler and mixed precision training sampler for our dataset. The fourth model, labeled as the "ResNet50 and More," introduces a more complex architecture by combining a ResNet50 backbone with additional fully connected layers. The training setup is similar to the CUDA-accelerated model, with data parallelism implemented using MPI to distribute the workload across multiple processes. Mixed precision training is also applied to enhance training efficiency by using lower-precision data types.

In all four cases, the code includes the necessary preprocessing steps, model definitions, training loops, and testing procedures, showcasing the flexibility and adaptability of the provided code for different models and training scenarios. The code also provides options for distributed training using MPI and mixed precision training for further optimization, however, we have yet to explore the effective use of these aforementioned strategies.

3. RESULTS

*H. X-Ray Image Disease Detection*

Disease Detection or Classification accuracy is computed by evaluating the model's predictions against the ground truth labels for the test dataset. The process begins with applying the model to the test images using the testing data loader, generating predicted probabilities for each label. To facilitate binary classification, a threshold of 0.5 is employed, categorizing predictions above this threshold as positive (1) and those below as negative (0). Subsequently, these binary predictions are compared with the ground truth labels for each image. For each label, correctness is determined by assessing whether the predicted value aligns with the ground truth value, and the count of correct predictions is tallied. The final step involves calculating accuracy by dividing the total count of correct predictions by the overall number of predictions, which is the product of the number of images and the total number of labels (14 classes). We can observe the evaluation accuracy of our different models and strategies in **Table 1**.

| Model | Training Strategy | Test Accuracy |
|---|---|---|
| ResNet18 | No accelerated strategy | 86.12% |
| ResNet18 | CUDA-accelerated | 86.26% |
| ResNet18 | CUDA + MPI | 86.29% |
| ResNet18 | CUDA + MPI + Mixed Precision Training + Learning Rate Scheduler | 86.34% |
| **ResNet50** | **CUDA + MPI + Mixed Precision Training + Learning Rate Scheduler** | **86.92%** |

*Table 1*: Test accuracy of different model training strategies. ResNet50 with accelerated training yielded a higher test accuracy of 86.92%.

*I. Training Execution Time*

The performance results obtained from our experiments demonstrate significant efficiency gains achieved through parallel implementations in model training tasks. The baseline ResNet18 model without any accelerated strategies exhibited the longest training time, amounting to 150.38 minutes. In contrast, the CUDA-accelerated ResNet18 model, leveraging the GPU for parallel processing, significantly reduced the training time to 10.19 minutes, showcasing the efficiency gained through GPU acceleration. The subsequent model, implementing CUDA alongside MP4Py (Message Passing Interface for Python), maintained a short training time of 10.44 minutes, suggesting that additional parallelism strategies can further enhance efficiency. The following configuration, which included CUDA, MPI4Py, Mixed Precision Training, and a Learning Rate Scheduler, resulted in a training time of 10.45 minutes. Finally, the ResNet50 model, incorporating CUDA, MPI4Py, Mixed Precision Training, and a Learning Rate Scheduler, exhibited a slightly longer training duration of 12.13 minutes compared to the ResNet18 counterparts, emphasizing the potential impact of model architecture on training times. Overall, these results highlight the effectiveness of GPU acceleration and parallel processing strategies in optimizing deep learning model training times as shown in **Table 2**.

| Model | Training Strategy | Execution Time |
|---|---|---|
| ResNet18 | No accelerated strategy | 150.38 minutes |
| **ResNet18** | **CUDA-accelerated** | **10.19 minutes** |
| ResNet18 | CUDA + MPI | 10.44 minutes |
| ResNet18 | CUDA + MPI + Mixed Precision Training + Learning Rate Scheduler | 10.45 minutes |
| ResNet50 | CUDA + MPI + Mixed Precision Training + Learning Rate Scheduler | 12.13 minutes |

*Table 2*: Execution time of different model training strategies. ResNet18 with CUDA-accelerated training yielded lower runtime of 10.19 minutes.

*J. Training Loss Convergence*

Binary Cross Entropy with Logits loss is employed to compute the loss during the training of models. The goal is to minimize this loss, which signifies the dissimilarity between the model's predictions (logits) and the ground truth labels. The process of loss minimization, often involving gradient descent, is fundamental in training machine learning models to make accurate predictions. We compare the training loss of our ResNet18 and our CUDA-accelerated ResNet18 in **Figure 3**.

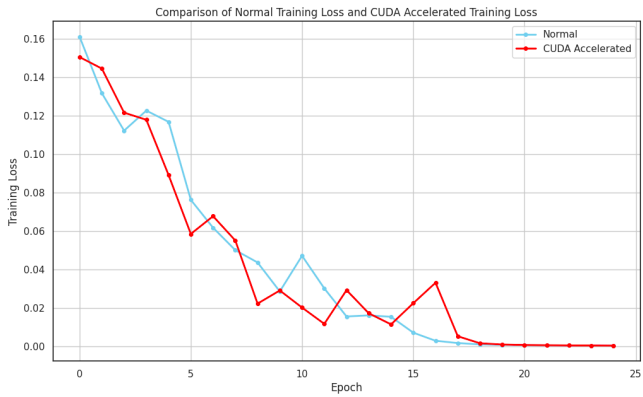

*Figure 3: Comparison of Normal Training Loss (in blue) and CUDA-accelerated Training Loss (in red)*

Moreover, all of our models convergence arbitrarily well over 25 epochs shown **Figures 4, 5, 6, and 7**.

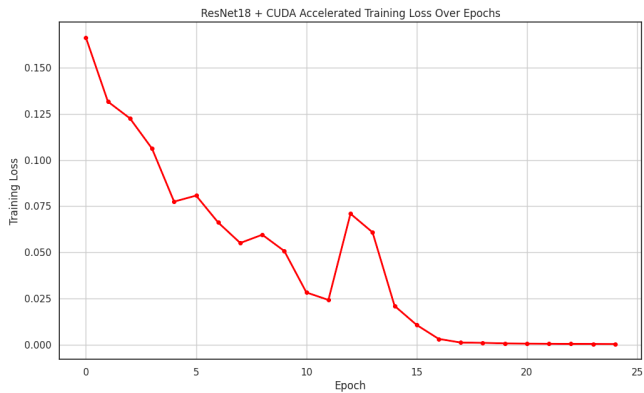

*Figure 4: Training loss for ResNet-18 with CUDA-accelerated training over Epochs*

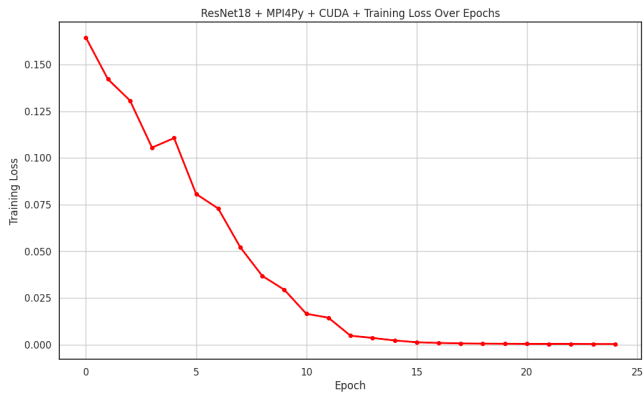

*Figure 5: Training loss for ResNet-18 with CUDA and MPI4Py training over Epochs*

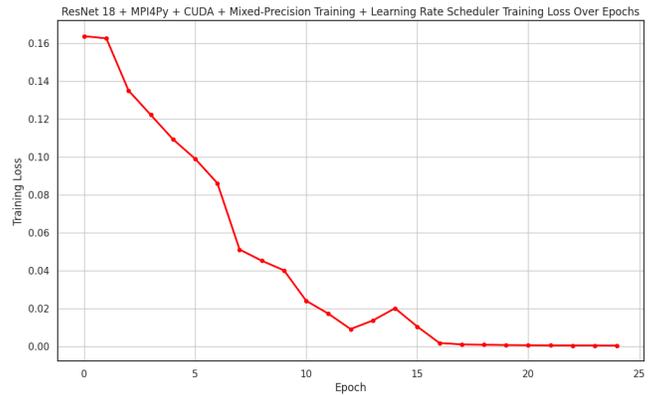

*Figure 6: Training loss for ResNet-18 with CUDA, MPI4Py, Mixed Precision Training, and Learning Rate Scheduler training over Epochs*

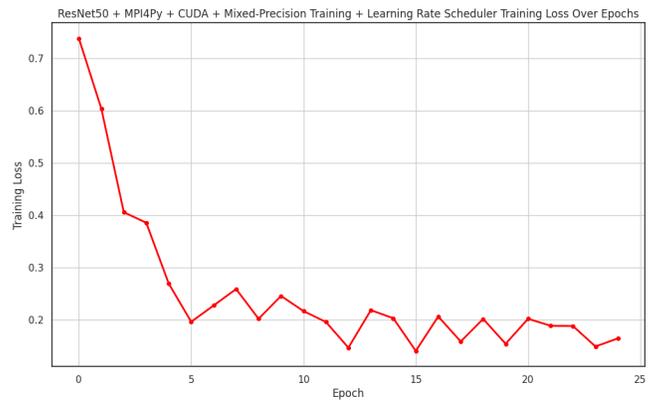

*Figure 7: Training loss for ResNet-18 with CUDA, MPI4Py, Mixed Precision Training, and Learning Rate Scheduler training over Epochs*

4. CONCLUSIONS

In this study, we center our project to address the challenge of X-ray image-based disease diagnosis and multi-class image classification models addresses a critical drawback related to the prolonged execution time of training these elaborate models. The study explores various strategies within the PyTorch framework, including CUDA acceleration, MPI4Py, and mixed precision training, to optimize training efficiency. The results showcase significant improvements in training times, with the baseline ResNet18 model taking 150.38 minutes, the CUDA-accelerated variant reducing it to 10.19 minutes, and further parallelization strategies maintaining efficient training times. The ResNet50 model, incorporating additional complexities, exhibited a slightly longer duration of 12.13 minutes. Accuracy evaluations demonstrate the effectiveness of the proposed strategies in achieving accurate disease detection.

Potential future work involves delving deeper into the explored strategies, particularly the untapped potential of MPI4Py and mixed precision training. Further optimization and fine-tuning of these parallelization methods may yield even more substantial efficiency gains. Additionally, investigating the scalability of these strategies with larger datasets and more complex models could provide insights into their adaptability to broader medical imaging applications. The study might also explore ensemble learning approaches, combining the strengths of different models to enhance overall performance. Addressing these avenues could contribute to the development of robust and efficient models for timely medical diagnosis through X-ray images, addressing the persistent challenge of computational efficiency in the medical image processing domain.